\documentclass[runningheads]{llncs}

\usepackage{eccv}

\usepackage{eccvabbrv}

\makeatletter
\newcommand{\makesupptitle}{%
  \let\orig@title\@title
  \title{Supplementary Material for Geometric Distillation from Rectified Stereo: Leveraging Epipolar Cues for Monocular Depth}%
  \author{\quad} % 시각적으로 보이지 않는 띄어쓰기 기호 입력
  \institute{\vspace{-4em}}
  
  \maketitle
}
\makeatother

\usepackage{graphicx}
\usepackage{booktabs}
\usepackage{multirow}
\usepackage{enumitem}
\usepackage{hhline}
\usepackage{wrapfig}
\usepackage{caption}
\usepackage{tabularx}
\usepackage{colortbl}
\usepackage[accsupp]{axessibility}  % Improves PDF readability for those with disabilities.

\usepackage{hyperref}
\usepackage{orcidlink}

\begin{document}

% ---------------------------------------------------------------
% TODO REVIEW: Replace with your title
\title{Geometric Distillation from Rectified Stereo: Leveraging Epipolar Cues for Monocular Depth} 

% TODO REVIEW: If the paper title is too long for the running head, you can set
% an abbreviated paper title here. If not, comment out.
\titlerunning{EpiDistill}

% TODO FINAL: Replace with your author list. 
% Include the authors' OCRID for the camera-ready version, if at all possible.
\author{Jung-Hee Kim\inst{1}\orcidlink{0009-0007-3198-3029} \and
Xiaoming Liu\inst{1,2}\orcidlink{0000-0003-3215-8753}}

% TODO FINAL: Replace with an abbreviated list of authors.
\authorrunning{J.~Kim and X.~Liu}
% First names are abbreviated in the running head.
% If there are more than two authors, 'et al.' is used.

% TODO FINAL: Replace with your institution list.
\institute{Michigan State University, East Lansing, MI 48824, USA \\ \and 
University of North Carolina at Chapel Hill, Chapel Hill, NC 27514, USA \\
\email{kimjun84@msu.edu, liuxm@cs.unc.edu}}

\maketitle
\begin{abstract}
Monocular depth foundation models have demonstrated remarkable generalization capabilities across diverse environments. However, they continue to struggle with metric depth estimation in diverse environments. This limitation stems from the inherent scale ambiguity of single-view inference, leading to misaligned scale predictions even when the relative geometry is accurate. Conversely, recent multi-view foundation models leverage cross-view cues to learn robust scene-level geometry and consistent scale. Yet, these benefits typically vanish during single-image inference, as the absence of explicit geometric constraints causes performance to degrade. To bridge this gap, we propose a novel framework that transfers the scale-aware geometric priors of multi-view models into monocular depth foundation models. Specifically, we introduce an Epipolar Distillation (EpiDistill), an approach utilizing Rectified Stereo Tokens, which enables the single-view prediction model to retain epipolar attention patterns and maintain geometric consistency without requiring multi-view inputs at inference. Experimental results demonstrate that our method significantly improves zero-shot metric depth estimation, particularly on challenging datasets like ETH3D and DIODE where scale alignment is critical. Furthermore, our approach is model-agnostic, consistently boosting the performance of state-of-the-art ViT-based models, including UniDepthV2 and DepthPro. \href{https://jungheekim29.github.io/EpiDistill/}{\textit{Project Link}}

\end{abstract}

\section{Introduction}
\label{sec:intro}

Monocular depth foundation models \cite{da1:yang2024depth, da2:yang2024depth, da3:lin2025depth, midas:birkl2023midas, unidepth:piccinelli2024unidepth, unidepthv2:piccinelli2025unidepthv2, depthpro:bochkovskii2024depth, addmonofound:guizilini2023towards} have demonstrated remarkable generalization capabilities across diverse environments. Despite being trained on massive datasets, these models inherently struggle to perceive the correct depth scale, a fundamental challenge in metric depth estimation. To resolve this scale ambiguity, early methods \cite{metric3d:yin2023metric3d, metric3dv2:hu2024metric3d} leveraged camera intrinsics at inference time to disentangle focal length from depth predictions. Recent approaches have relaxed this constraint by directly regressing camera intrinsics alongside depth \cite{moge1:wang2025moge, moge2:wangmoge, unidepth:piccinelli2024unidepth, unidepthv2:piccinelli2025unidepthv2}, relying on large-scale datasets to implicitly learn the correlation between the two.

Concurrently, the recent advent of multi-view foundation models \cite{vggt:wang2025vggt, da3:lin2025depth, ma:keetha2025mapanything} has enabled robust, scene-level 3D reconstruction from a sequence of images. These methods exploit task-specific tokens and leverage geometry-aware learning through global cross-attention layers. By processing multi-view inputs in a single feed-forward pass, they successfully establish geometric correspondence and predict a globally consistent metric scale. 

However, a critical limitation arises when these multi-view foundation models are tasked with single-view inference. Restricted to single-view inputs, their performance degrades significantly, often falling short of dedicated monocular models. This degradation fundamentally stems from an architectural collapse: in the absence of multi-view input, the cross-view global attention degenerates into a standard intra-frame self-attention. This creates a geometry mismatch between training and inference, leading to the immediate loss of the scale-aware geometric priors learned from multi-view data. 
\begin{figure}[t]
    \centering
    \includegraphics[width=1.0\linewidth]{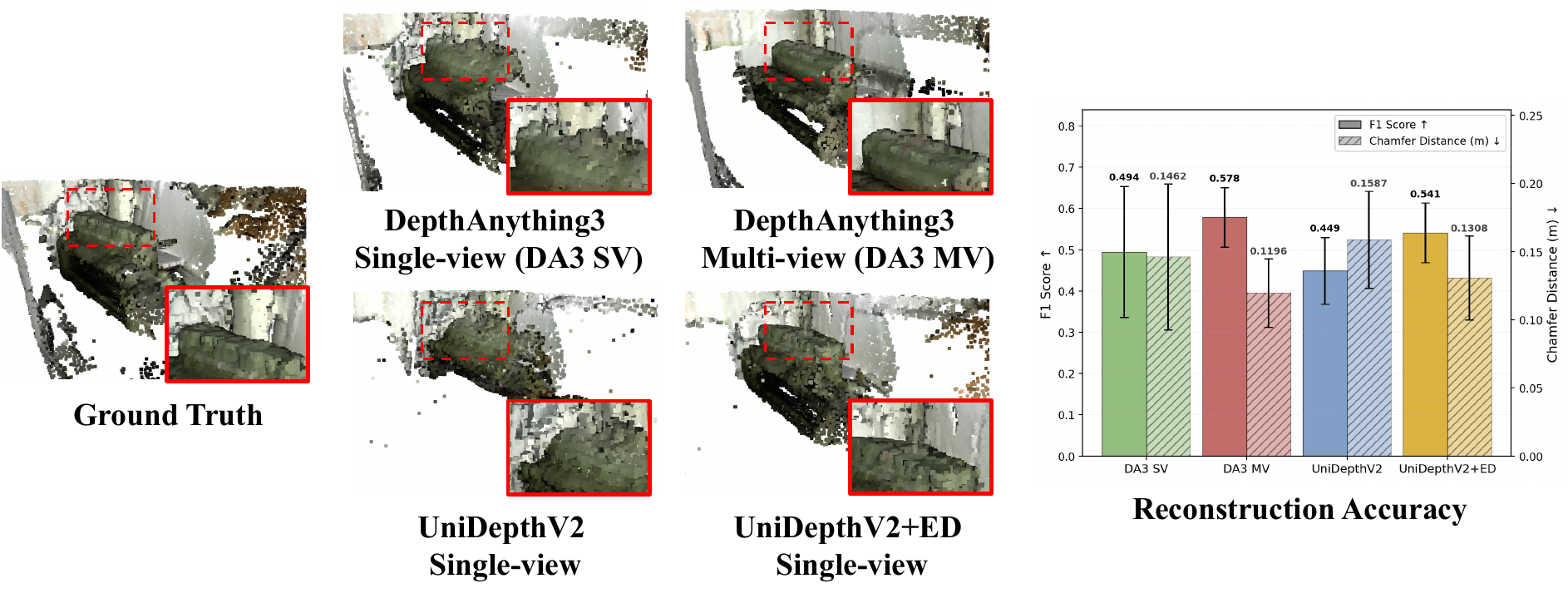}
    \caption{\textbf{3D Scene Reconstruction on ScanNet++ \cite{scannet++:yeshwanth2023scannet++}.} \textbf{(Left)} Qualitative comparison of point clouds accumulated across multiple sequential frames using single-view (SV) and multi-view (MV) methods. \textbf{(Right)} Quantitative evaluation using F1 Score ($\uparrow$) and Chamfer Distance ($\downarrow$) on the test set. ED indicates the proposed \textit{EpiDistill} method.}
    \label{fig:limit_pcl}
\end{figure}

We empirically demonstrate this limitation in \cref{fig:limit_pcl}. Given multi-view images with camera parameters, a multi-view model such as DepthAnything3 (DA3) \cite{da3:lin2025depth} explicitly processes 5-frame batches to align the scale, generating well-aligned point clouds in a single feed-forward pass. Conversely, when forced to perform single-view inference and accumulate results sequentially, the model severely struggles with per-frame scale ambiguity, producing misaligned reconstructions. This confirms that vital geometric structures, particularly metric scale, are inherently lost when the multi-view attention mechanism collapses. To date, how to effectively project the scale robustness of multi-view models into single-view prediction remains an open challenge.

To bridge this gap, we introduce \textit{EpiDistill}, a novel geometric distillation framework that explicitly transfers multi-view geometric knowledge to a single-view model. During training, a multi-view model achieves accurate depth perception by utilizing an explicit \textit{Depth-Guided Epipolar Attention}. To process single-view inputs while preserving this learned structure, our single-view model pairs the retained epipolar attention with learnable \textit{Rectified Stereo Tokens}, directly distilling knowledge from the multi-view tokens. These tokens serve as a structural anchor to provide the geometric guidance necessary to effectively leverage the distilled priors. As visually and quantitatively evidenced in \cref{fig:limit_pcl}, integrating our method with a baseline, UniDepthV2+\textit{EpiDistill} (ED), successfully produces accurate, globally scale-aligned reconstructions from single-view inputs—improving the F1 score by 20.5\% and reducing the Chamfer distance by 17.6\%. By inheriting the scale robustness inherent to multi-view geometry, our approach significantly elevates the zero-shot metric depth estimation capabilities of state-of-the-art (SoTA) Vision Transformer (ViT) \cite{vit:dosovitskiy2020image} baselines. Our main contributions are summarized as follows:

\begin{itemize}
\item We propose \textit{EpiDistill}, a novel framework that transfers scale-robust multi-view geometric priors to single-view prediction models.
\item We design a \textit{Depth-Guided Epipolar Attention} that utilizes ground-truth spatial bias along the epipolar line, providing explicit geometric supervision during the multi-view training phase.
\item We introduce \textit{Rectified Stereo Tokens}, which prevent the structural collapse of cross-view attention during single-view inference by serving as a geometric anchor in a rectified stereo setup.
\item Extensive experiments demonstrate that our approach is model-agnostic and significantly improves the zero-shot metric depth performance of SoTA ViT-based models (\textit{e.g.}, UniDepthV2, DepthPro), particularly on challenging datasets demanding strict scale alignment.
\end{itemize}

\section{Related Works}
\subsection{Monocular Depth Estimation}
Monocular depth estimation \cite{convdepth:laina2016deeper, depthex:xu2017multi, eigendepth:eigen2014depth} aims to predict the depth of a scene from a single input image. Early deep learning approaches incorporated depth priors to improve performance, utilizing techniques such as ordinal regression \cite{orddep:fu2018deep} and adaptive binning to partition the depth range \cite{adabin:bhat2021adabins}. However, these models heavily relied on direct ground-truth supervision, which fundamentally limited their generalization capabilities across diverse datasets. To mitigate the reliance on annotated data, self-supervised methods \cite{unsup1:choi2021adaptive, ddad:guizilini20203d, unsup3:poggi2020uncertainty, unsup4:watson2019self, unsup5:wong2020unsupervised, SS-SFM:Revisit-self-supervision-with-local-structure-from-motion} were proposed. These approaches leverage novel view synthesis, utilizing nearby video frames and photometric consistency losses to jointly estimate depth and camera pose without explicit ground-truth depth. 

Recently, depth foundation models trained on massive, diverse datasets have emerged. By adopting robust depth representations and range normalization techniques \cite{da1:yang2024depth, da2:yang2024depth, midas:birkl2023midas, dpt:ranftl2021vision}, these models successfully aggregate training data from various domains to achieve remarkable zero-shot relative depth estimation. Building upon this success, several recent works have focused on metric depth estimation \cite{unidepth:piccinelli2024unidepth, unidepthv2:piccinelli2025unidepthv2, depthpro:bochkovskii2024depth, unik3d:piccinelli2025unik3d, unidac:unidac-universal-metric-depth-estimation-for-any-camera} by employing additional modules to predict absolute scale. Nevertheless, despite leveraging large-scale datasets, monocular metric depth estimation models inherently suffer from scale ambiguity due to their training relying on limited 2D visual cues. In contrast, our approach resolves this ambiguity by supervising the model with explicit multi-view geometric constraints during training. By distilling the capacity to reason over epipolar geometry into a single-view model via rectified stereo tokens, our framework accurately predicts metric scale during single-view inference.

\begin{figure}[t]
    \centering
    \includegraphics[width=1.0\linewidth]{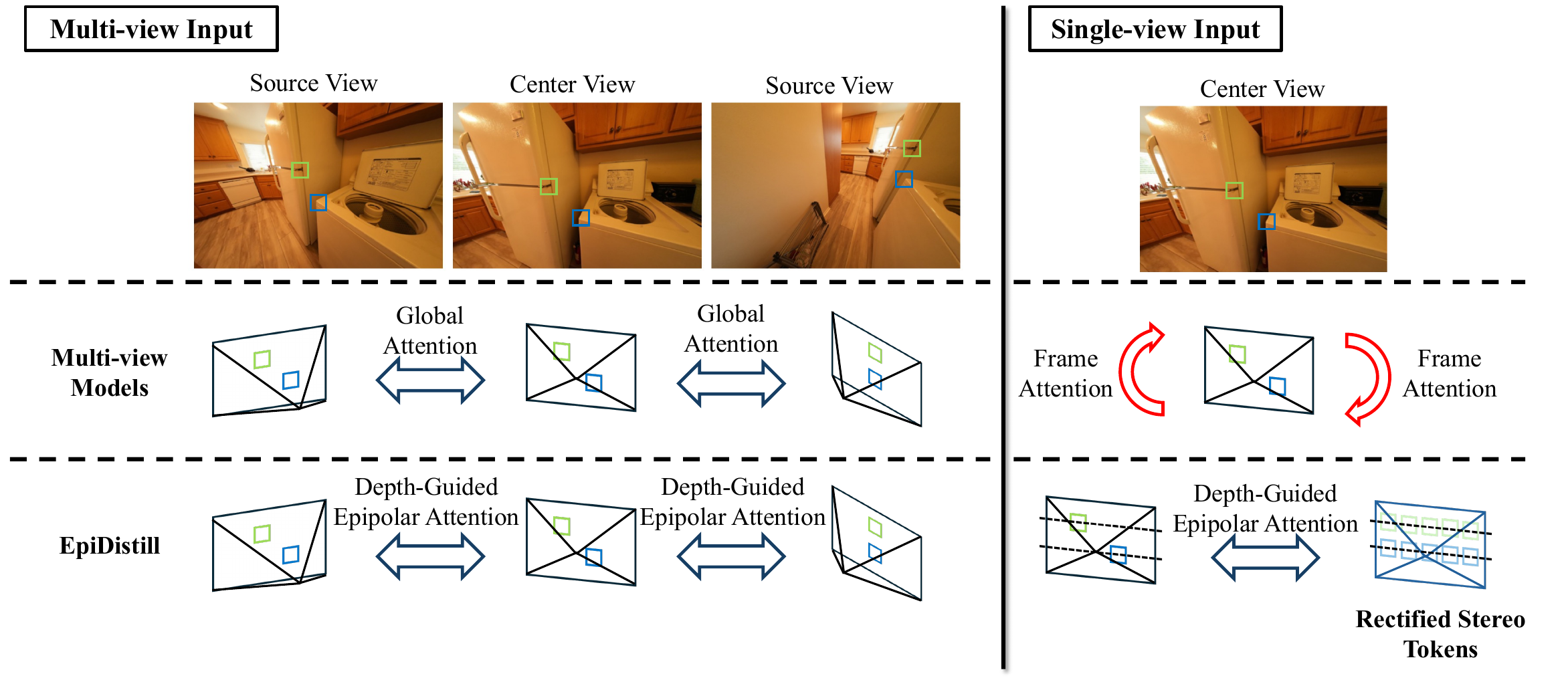}
    \caption{\textbf{Geometric ambiguity in single-view inference of multi-view models.} In standard architectures, the global attention layer collapses into standard intra-frame self-attention when deprived of auxiliary views. In contrast, our proposed \textit{EpiDistill} framework preserves the structural integrity of the cross-view attention pathways by utilizing depth-guided epipolar attention coupled with rectified stereo tokens.}
    \label{fig:mvf_limit}
\end{figure}

\subsection{Multi-view Foundation Models}
Recent advances in 3D vision have introduced multi-view foundation models such as DUSt3R \cite{dust3r:wang2024dust3r} and MASt3R \cite{mast3r:leroy2024grounding}, which utilize multiple viewpoints to jointly predict dense geometry, including point clouds and camera information. However, these methods strictly require two views as input and rely on computationally expensive iterative processes for global scene reconstruction. To overcome these computational bottlenecks, models such as VGGT \cite{vggt:wang2025vggt}, DepthAnything3 \cite{da3:lin2025depth}, and MapAnything \cite{ma:keetha2025mapanything} are proposed. They achieve globally aligned scene reconstruction and robust scale estimation through a single feed-forward pass using multi-view inputs. These architectures utilize alternating token mixing strategies, specifically interleaving intra-frame (frame) and cross-frame (global) attention layers, to establish multi-view consistency and infer scene-level scale. 

Consequently, a limitation arises during single-view inference: the global attention naturally degenerates into frame attention, as shown in \cref{fig:mvf_limit}. This structural collapse discards the rich, scale-aware geometric priors learned during multi-view training, resulting in scale ambiguity similar to traditional monocular models, as shown in \cref{fig:limit_pcl}. Unlike these approaches, our proposed \textit{EpiDistill} framework explicitly bridges this gap. By guiding the monocular prediction model to retain the epipolar constraints learned from multi-view training, we successfully preserve robust scale estimation capabilities without requiring additional views at inference time.

\section{Preliminaries}
In this section, we revisit the architecture of multi-view foundation models \cite{vggt:wang2025vggt,da3:lin2025depth, ma:keetha2025mapanything}, which serves as the structural basis for our approach. Specifically, we formalize the tokenization and attention mechanisms responsible for learning geometric scale priors.

\noindent \textbf{Multi-View Tokenization.} Given a set of $N$ images $\{\mathbf{I}_1, \mathbf{I}_2, \dots, \mathbf{I}_N \}$, a Vision Transformer (ViT) \cite{vit:dosovitskiy2020image} encoder processes each image into a sequence of spatial tokens. Let $\mathbf{F}_i \in \mathbb{R}^{L \times d}$ denote the spatial tokens for image $\mathbf{F}_i$, where $L$ is the sequence length and $d$ is the feature dimension. To explicitly reason about camera geometry and global scale, learnable task-specific tokens (\textit{e.g.}, camera or scale) are appended to the spatial token sequence of each view.

\noindent \textbf{Frame and Global Attention.} The  representational power of recent multi-view foundation models \cite{vggt:wang2025vggt, da3:lin2025depth, ma:keetha2025mapanything} stems from alternating attention blocks. The frame attention operates strictly within individual views, applying standard self-attention to $\mathbf{F}_i$. 
Conversely, the global (cross-view) attention models broad spatial relationships and structural priors across different viewpoints. Let $\mathbf{F}_{\text{global}} = [\mathbf{F}_1; \mathbf{F}_2; \dots; \mathbf{F}_N] \in \mathbb{R}^{(N \cdot L) \times d}$ be the concatenated token sequence from all views. The global attention computes the output features by allowing tokens to attend to tokens derived from all viewpoints. With this cross-view interaction, the network inherently learns to resolve scale ambiguity by extracting multi-view geometric constraints. The final decoded outputs yield geometry predictions including depth maps and camera intrinsics.

\begin{figure}[t]
    \centering
    \includegraphics[width=1.0\linewidth]{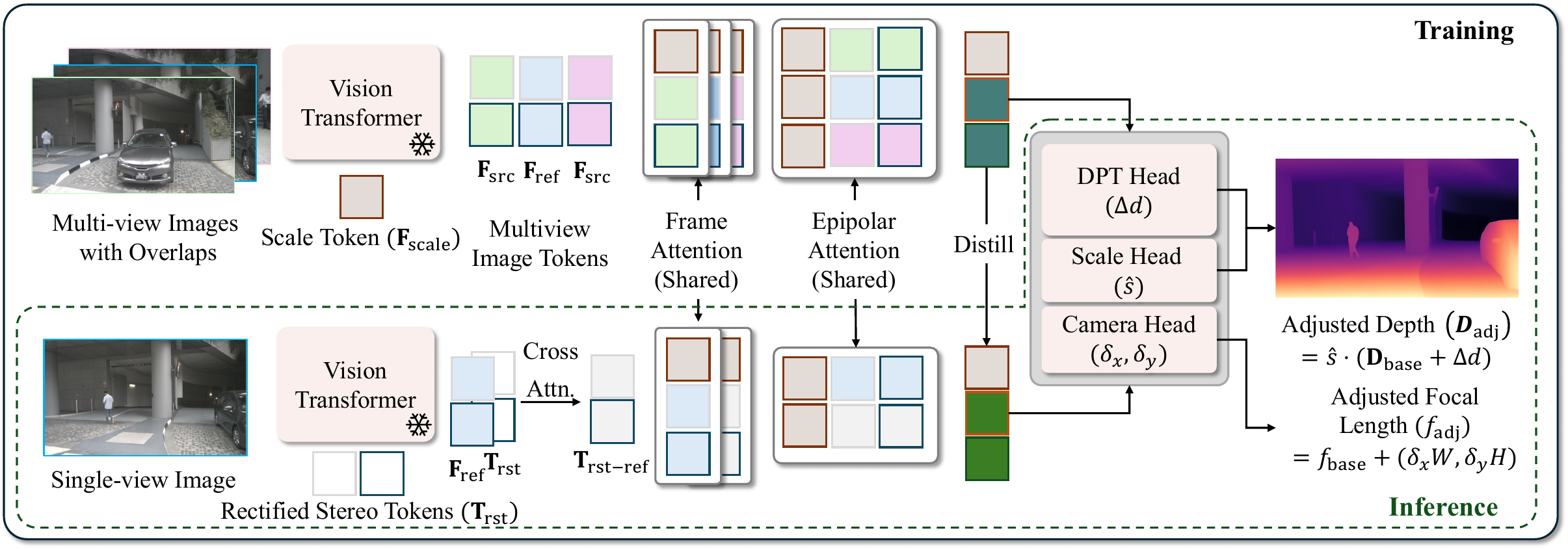}
    \caption{\textbf{Overview of the EpiDistill architecture.} \textbf{(Top)} During training, a multi-view model extracts geometric priors via frame attention and depth-guided epipolar attention. \textbf{(Bottom)} A single-view model employs the shared layers but leverages \textit{Rectified Stereo Tokens} as geometric guidance. Knowledge is distilled from multi-view to single-view tokens. Finally, lightweight heads predict structural shift ($\Delta d$), global scale ($\hat{s}$), and focal length ($\delta_x, \delta_y$) adjustments to refine the baseline prediction. At inference, only the single-view pipeline is utilized.}  
    \label{fig:method}
    % \vspace{-5pt}
\end{figure}

\section{EpiDistill}

The overall pipeline of our proposed framework is illustrated in \cref{fig:method}. Our approach leverages multi-view image sets (\textit{e.g.}, $\mathbf{I}_{\text{ref}}, \mathbf{I}_{\text{src}}, \dots$) during training to extract geometrically consistent scale priors. Specifically, we build upon SoTA ViT-based \cite{vit:dosovitskiy2020image} depth foundation models, such as UniDepthV2 \cite{unidepthv2:piccinelli2025unidepthv2} and DepthPro \cite{depthpro:bochkovskii2024depth}. Given overlapping multi-view images, a frozen model encoder extracts dense spatial tokens ($\mathbf{F}_\text{ref}, \; \mathbf{F}_\text{src}, \dots$). We append a learnable scale token ($\mathbf{F}_\text{scale}$) to the spatial tokens of each individual view and process them through shared frame attention layers. Subsequently, the epipolar attention is applied across views to explicitly learn multi-view geometric correspondences and resolve scale ambiguity.

To enable highly accurate single-view prediction, we distill this consistent geometric information into a single-view model at the token level. To preserve the learned epipolar attention without requiring actual multi-view inputs, the single-view model substitutes the missing source views with a set of learnable \textit{Rectified Stereo Tokens} ($\mathbf{T}_\text{rst}$). These tokens interact with the single-view image features via a cross-attention layer to form reference-guided tokens ($\mathbf{T}_\text{rst-ref}$). Reference-guided tokens are then concatenated with the primary spatial tokens and processed through the shared epipolar attention layers. Finally, the geometric knowledge from the multi-view attended tokens is explicitly distilled into these single-view embeddings, yielding attention-refined spatial tokens.

Because the underlying baseline models already excel at relative depth estimation, we explicitly decouple the final depth prediction into structural and scale components. Given the original depth prediction from the foundation model as $\mathbf{D}_{\text{orig}}$, we retain this dense prediction and normalize it to obtain a scale-invariant structural depth map, $\mathbf{D}_{\text{base}} = \mathbf{D}_{\text{orig}} / \text{median}(\mathbf{D}_{\text{orig}})$, where $N$ is the total number of pixels. The attention-refined spatial tokens are then fed into a lightweight Dense Prediction Transformer (DPT) \cite{dpt:ranftl2021vision} head to predict a dense structural shift offset, $\Delta d$. Concurrently, the scale token, enriched by both frame and epipolar attention, is passed through an MLP-based scale head to predict a metric scale factor, $\hat{s}$. The final adjusted metric depth, $\mathbf{D}_{\text{adj}}$, explicitly fuses these decoupled outputs:
\begin{equation}
    \mathbf{D}_{\text{adj}} = \hat{s} \cdot (\mathbf{D}_{\text{base}} + \Delta d).
\end{equation}
An auxiliary camera head resolves camera and scale ambiguities by predicting normalized focal length residuals $(\delta_x, \delta_y)$ from scale tokens. These are rescaled by $(W, H)$ to compute the adjusted focal length: $f_{\text{adj}} = f_{\text{base}} + (\delta_x W, \delta_y H)$.

\begin{figure}[t]
    \centering
    \includegraphics[width=0.9\linewidth]{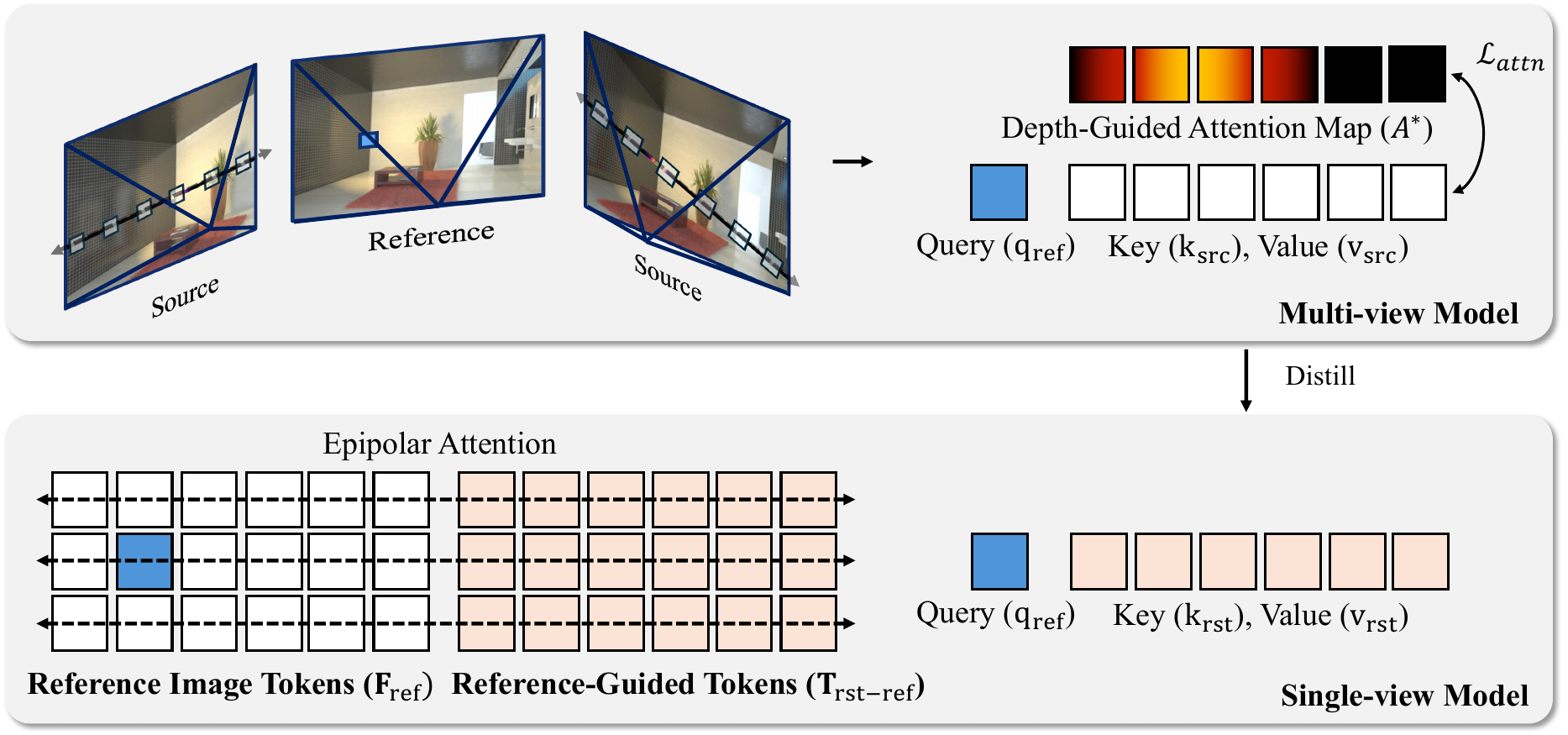}
    \caption{\textbf{Depth-Guided Epipolar Attention and Rectified Stereo Tokens.} \textbf{(Top)} In the multi-view model, a reference query (blue) attends to points along the 2D epipolar line, supervised by $\mathcal{L}_{\text{attn}}$ to localize 3D correspondences. \textbf{(Bottom)} In  single-view model, reference-guided tokens ($\text{T}_\text{rst-ref}$) substitute source views, simplifying the geometry into a 1D horizontal search space (dashed arrows).}
    \label{fig:rstoken}
\end{figure}

\subsection{Depth-Guided Epipolar Attention}
\label{epipolar}
We propose a novel epipolar attention network designed to effectively fuse multi-view observations and extract robust geometric priors. For a given pair of images, depth-guided epipolar attention computes cross-attention between the tokens of the reference view and those of the source view. To enforce geometric consistency, the attention for a query token from the reference image is constrained to tokens sampled along its corresponding epipolar line in the source image.

Let $\mathbf{K}_{\text{ref}}$ and $\mathbf{K}_{\text{src}}$ be the intrinsic matrices of the reference and source view, respectively, and $[\mathbf{R}| \mathbf{t}]$ be the relative rigid transformation between the reference view and the source view. For a query pixel $\mathbf{p}_{\text{ref}, i}$ in the reference view (where $i \in \{1, \dots, N\}$ indexes the total $N$ reference tokens) with homogeneous coordinates $\tilde{\mathbf{p}}_{\text{ref}, i}$, the corresponding epipolar line $\mathbf{l}_{\text{src}, i}$ in the source view is defined by the fundamental matrix $\mathbf{F} \in \mathbb{R}^{3 \times 3}$:
\begin{equation}
\mathbf{l}_{\text{src}, i} = \mathbf{F} \tilde{\mathbf{p}}_{\text{ref}, i} = \mathbf{K}_{\text{src}}^{-\top} [\mathbf{t}]_{\times} \mathbf{R} \mathbf{K}_{\text{ref}}^{-1} \tilde{\mathbf{p}}_{\text{ref}, i},
\end{equation}
where $[\mathbf{t}]_{\times}$ is the skew-symmetric matrix of the translation vector $\mathbf{t}$. While standard cross-attention is applied over a set of sampled points along this line, we introduce an explicit geometric guidance mechanism using ground-truth depth during training. As illustrated in \cref{fig:rstoken}, by leveraging the ground-truth depth $\mathbf{D}_{\text{gt}}(\mathbf{p}_{\text{ref}, i})$, the exact corresponding target point $\mathbf{p}^*_{\text{src}, i}$ in the source view can be determined by back-projecting the reference pixel to 3D and projecting it onto the source image plane:
\begin{equation}
\mathbf{p}^*_{\text{src}, i} = \pi \left( \mathbf{K}_{\text{src}} \left( \mathbf{R} \left( \mathbf{D}_{\text{gt}}(\mathbf{p}_{\text{ref}, i}) \mathbf{K}_{\text{ref}}^{-1} \tilde{\mathbf{p}}_{\text{ref}, i} \right) + \mathbf{t} \right) \right),
\end{equation}
where $\pi([x, y, z]^\top) = [x/z, y/z]^\top$ denotes the perspective projection operation.

To concentrate attention near the exact correspondence, we introduce a spatial Gaussian bias $B_{ij}$ for each sampled coordinate $\mathbf{p}_{\text{src}, ij}$, where $j \in \{1, \dots, M\}$ indexes the $M$ discrete points sampled along the epipolar line $\mathbf{l}_{\text{src}, i}$:
\begin{equation}
B_{ij} = -\gamma \| \mathbf{p}_{\text{src}, ij} - \mathbf{p}^*_{\text{src}, i} \|^2_2,
\end{equation}
where $\gamma$ controls the distribution sharpness (\textit{e.g.,} $\gamma=50$). This bias is added directly to the raw cross-attention logits $E_{ij} = \mathbf{q}_{\text{ref}, i}^\top \mathbf{k}_{\text{src}, j} / \sqrt{d}$, yielding the depth-guided target attention map $A^*_{ij} = \operatorname{Softmax}_j(E_{ij} + B_{ij})$. Here, $\mathbf{q}_{\text{ref}, i}$ and $\mathbf{k}_{\text{src}, j}$ denote the query token from the reference view and the key token from the source view, respectively. To distill this capability, we apply a cross-entropy loss over the predicted attention weights:
\begin{equation}
\mathcal{L}_{\text{attn}} = - \frac{1}{N} \sum_{i=1}^{N} \sum_{j=1}^{M} A^*_{ij} \log(\hat{A}_{ij}),
\label{eq:l_attn}
\end{equation}
where $\hat{A}_{ij} = \operatorname{Softmax}_j(E_{ij})$ represents the unguided attention predicted by the model. This aligns the attention distributions within the epipolar sampling space.
Note that $\mathbf{A}^*$ is detached from the gradient flow to serve solely as a supervision signal. Additionally, $\mathcal{L}_{\text{attn}}$ is masked out for invalid regions due to the sparsity of the ground-truth depth, while the epipolar attention remains fully active.

\subsection{Rectified Stereo Tokens}
\label{sec:rstoken}

In multi-view geometry, a rectified stereo setup provides an ideal constrained environment for depth perception. By aligning the image planes, the relative camera pose simplifies to an identity rotation ($\mathbf{R} = \mathbf{I}$) and a purely horizontal translation $\mathbf{t} = [b, 0, 0]^\top$. Assuming identical camera intrinsics with focal length $f$, the fundamental matrix $\mathbf{F}$ becomes:
\begin{equation}
    \mathbf{F} = \mathbf{K}_{\text{src}}^{-\top} [\mathbf{t}]_{\times} \mathbf{R} \mathbf{K}_{\text{ref}}^{-1} = \begin{bmatrix} 0 & 0 & 0 \\ 0 & 0 & -c \\ 0 & c & 0 \end{bmatrix},
\end{equation}
where $c = b/f$. For a given reference pixel $\tilde{\mathbf{p}}_{\text{ref}} = [x_0, y_0, 1]^\top$, the corresponding epipolar line in the source view is calculated as $\mathbf{l}_{\text{src}} = \mathbf{F} \tilde{\mathbf{p}}_{\text{ref}} = [0, -c, c y_0]^\top$. Since the baseline-dependent constant $c$ cancels out in the resulting 2D line equation ($-cy + cy_0 = 0$), the geometric search space is strictly confined to $y = y_0$. 

\begin{figure}[t]
    \centering
    \includegraphics[width=1.0\linewidth]{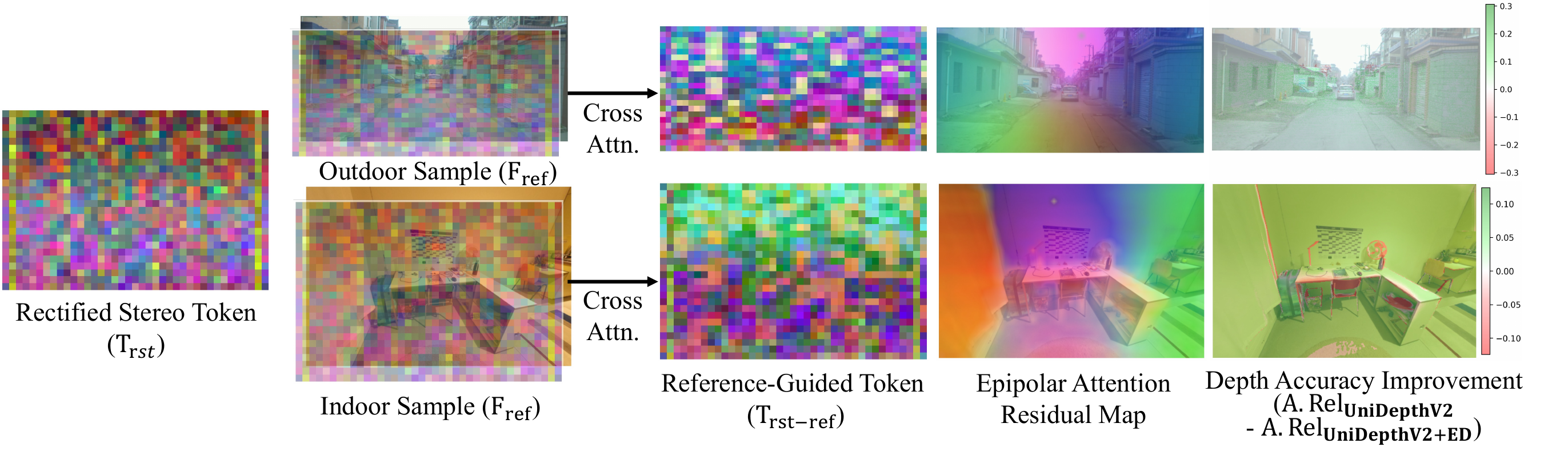}
    \caption{\textbf{Visualization of Rectified Stereo Tokens via Principal Component Analysis (PCA).} The visual progression shows the PCA-projected RGB mappings of the learnable spatial tokens ($\mathbf{T}_\text{rst}$) undergoing cross-attention with the encoded reference features ($\mathbf{F}_{\text{ref}}$) to generate reference-guided tokens ($\mathbf{T}_{\text{rst-ref}}$). Additionally, we visualize the PCA-reduced epipolar attention residual map, alongside the resulting depth accuracy improvement map, where green regions indicate reduced A.Rel error and red indicates degradation.}
    \label{fig:rstoken_detail}
    \vspace{-3mm}
\end{figure}

This baseline-agnostic property simplifies the epipolar attention, requiring the model to attend only to tokens along the same horizontal scanline. To preserve this structure without actual multi-view inputs, we introduce learnable \textit{Rectified Stereo Tokens} ($\mathbf{T}_{\text{rst}}$) to serve as a constant geometric anchor. We initialize a spatial token grid $\mathbf{T}_{\text{rst}} \in \mathbb{R}^{HW \times d}$ and dynamically interpolate it to match the spatial resolution of the encoded reference features $\mathbf{F}_{\text{ref}} \in \mathbb{R}^{H' W' \times d}$. Through cross-attention, $\mathbf{T}_{\text{rst}}$ queries $\mathbf{F}_{\text{ref}}$ to embed spatial context, yielding reference-guided tokens $\mathbf{T}_{\text{rst-ref}}$. Subsequently, the model applies the rectified epipolar attention between $\mathbf{F}_\text{ref}$ and $\mathbf{T}_{rst-ref}$, constraining the cross-attention interaction strictly along the horizontal scanlines to explicitly simulate the rectified stereo setup. 

To interpret this transformation, we visualize the high-dimensional internal representations—specifically the tokens ($\mathbf{T}_{\text{rst}}$, $\mathbf{F}_{\text{ref}}$, $\mathbf{T}_{\text{rst-ref}}$) and the epipolar attention residual map—by reducing their $d$-dimensional features to three principal components via PCA and mapping them to RGB channels. As shown in \cref{fig:rstoken_detail}, the cross-attention injects scene-specific structural awareness into $\mathbf{T}_{\text{rst-ref}}$. Furthermore, we analyze the PCA-mapped epipolar attention residual, defined as the feature difference before and after applying the epipolar attention layer. This residual demonstrates that the network successfully captures semantic information, such as scene layouts. By leveraging this semantically aware geometric anchor, \textit{EpiDistill} (ED) effectively guides the depth refinement process. This results in a significant reduction in the absolute relative error (A.Rel), as evidenced by the separate depth improvement map ($\text{A.Rel}_{\text{UniDepthV2}} - \text{A.Rel}_{\text{UniDepthV2+ED}}$), achieving accurate metric results.

\subsection{Loss Functions}
\label{subsec:loss} 

Our model is trained using a composite loss function that explicitly decouples structure and scale while enforcing multi-view consistency and stabilizing the distillation process. The total loss $\mathcal{L}_{\text{total}}$ is formulated as a weighted sum of its individual components:
\begin{equation}
\mathcal{L}_{\text{total}} = \mathcal{L}_{\text{rel}} + \mathcal{L}_{\text{scale}} + \lambda_{\text{ray}}\mathcal{L}_{\text{ray}} + \lambda_{\text{distill}}\mathcal{L}_{\text{distill}} + \lambda_{\text{attn}}\mathcal{L}_{\text{attn}},
\end{equation}
where the loss weights are empirically set to $\lambda_{\text{ray}}=0.3$, $\lambda_{\text{distill}}=1.0$, and $\lambda_{\text{attn}}=0.1$.

\noindent \textbf{Relative Depth Loss ($\mathcal{L}_{\text{rel}}$).} 
To stably learn metric scale in a multi-view setting, we explicitly decouple the depth estimation into a normalized structural component and a global scale component. For the structural target, the ground-truth depth $\mathbf{D}_\text{GT}$ is normalized by its median within the valid mask region $\mathcal{M}$. The predicted relative depth $\mathbf{\tilde{D}}_{\text{rel}}$ is formed by adding a structural offset $\Delta d$ to the base depth $\mathbf{D}_\text{base}$:
\begin{equation}
\mathbf{\tilde{D}}_{\text{rel}} = \mathbf{D}_{\text{base}} + \Delta d, \quad
\mathbf{\bar{D}}_{\text{GT}} = \frac{\mathbf{D}_{\text{GT}}}{\mathop{\text{median}}\limits_{p \in \mathcal{M}} (\mathbf{D}_{\text{GT}, p})}.
\end{equation}
The structural depth loss $\mathcal{L}_\text{depth}$ operates in log-space using an $L_1$ loss:
\begin{equation}
    \mathcal{L}_\text{depth} = \sqrt{\frac{1}{|\mathcal{M}|} \sum_{p \in \mathcal{M}} \bigl| \log \mathbf{\tilde{D}}_{\text{rel}, p} - \log \mathbf{\bar{D}}_{\text{GT}, p} \bigr|}.
\end{equation}
Following \cite{unidepthv2:piccinelli2025unidepthv2}, we incorporate an edge-aware loss $\mathcal{L}_\text{edge}$ using Sobel-filtered image gradients to preserve sharp depth discontinuities around strong object boundaries. The final relative depth loss is formulated as $\mathcal{L}_{\text{rel}} = \mathcal{L}_\text{depth} + \mathcal{L}_\text{edge}$.

\noindent \textbf{Scale Loss ($\mathcal{L}_\text{scale}$).}
The metric scale $\hat{s}$ predicted by the scale head is independently supervised by the arithmetic mean of the ground-truth depth. To ensure gradient flows to the scale head, we apply an $L_1$ penalty in log space:
\begin{equation}
    \mathcal{L}_\text{scale} = \sqrt{\Bigl| \log \hat{s} - \log \Bigl( \frac{1}{|\mathcal{M}|} \sum_{p \in \mathcal{M}} \mathbf{D}_{\text{GT}, p} \Bigr) \Bigr|}.
\end{equation}

\noindent \textbf{Ray/Intrinsic Loss ($\mathcal{L}_\text{ray}$).}
To jointly optimize camera intrinsics, we enforce an $L_1$ penalty on the predicted focal length residuals ($\delta_x, \delta_y$) relative to the ground-truth focal lengths normalized by image width $W$ and height $H$:
\begin{equation}
    \mathcal{L}_\text{ray} = \Bigl| \frac{f_{x, \text{base}}}{W}+ \delta_x - \frac{f_{x,\text{GT}}}{W} \Bigr| + \Bigl|  \frac{f_{y, \text{base}}}{H}+ \delta_y - \frac{f_{y,\text{GT}}}{H} \Bigr|.
\end{equation}

\noindent \textbf{Distillation Loss ($\mathcal{L}_\text{distill}$).}
To ensure stable knowledge transfer, this mechanism aligns single-view tokens $\mathbf{T}_s$ with multi-view epipolar-aggregated features $\mathbf{T}_m$ guided by the rectified stereo tokens. The loss combines a cosine similarity direction objective with a magnitude penalty over the valid token mask $\mathcal{N}$:
\begin{equation}
    \mathcal{L}_\text{distill} = \frac{1}{|\mathcal{N}|} \sum_{k \in \mathcal{N}} \Bigl( 1 - \cos\bigl(\mathbf{T}_{m, k}, \mathbf{T}_{s, k}\bigr) + 0.1 \bigl| \|\mathbf{T}_{m, k}\|_2 - \|\mathbf{T}_{s, k}\|_2 \bigr| \Bigr), 
\end{equation}
where $\mathcal{N}$ represents the valid token mask established by the epipolar attention, ensuring that feature alignment is focused on valid cross-view correspondences.

\noindent \textbf{Attention Loss ($\mathcal{L}_\text{attn}$).}
Lastly, the attention loss $\mathcal{L}_{\text{attn}}$ explicitly guides the epipolar correspondence via cross-entropy optimization over the predicted attention weights, as previously formulated in \cref{eq:l_attn}.

\section{Experiments}
\label{sec:exp}

\begin{figure}[t]
    \centering
    \includegraphics[width=1.0\linewidth]{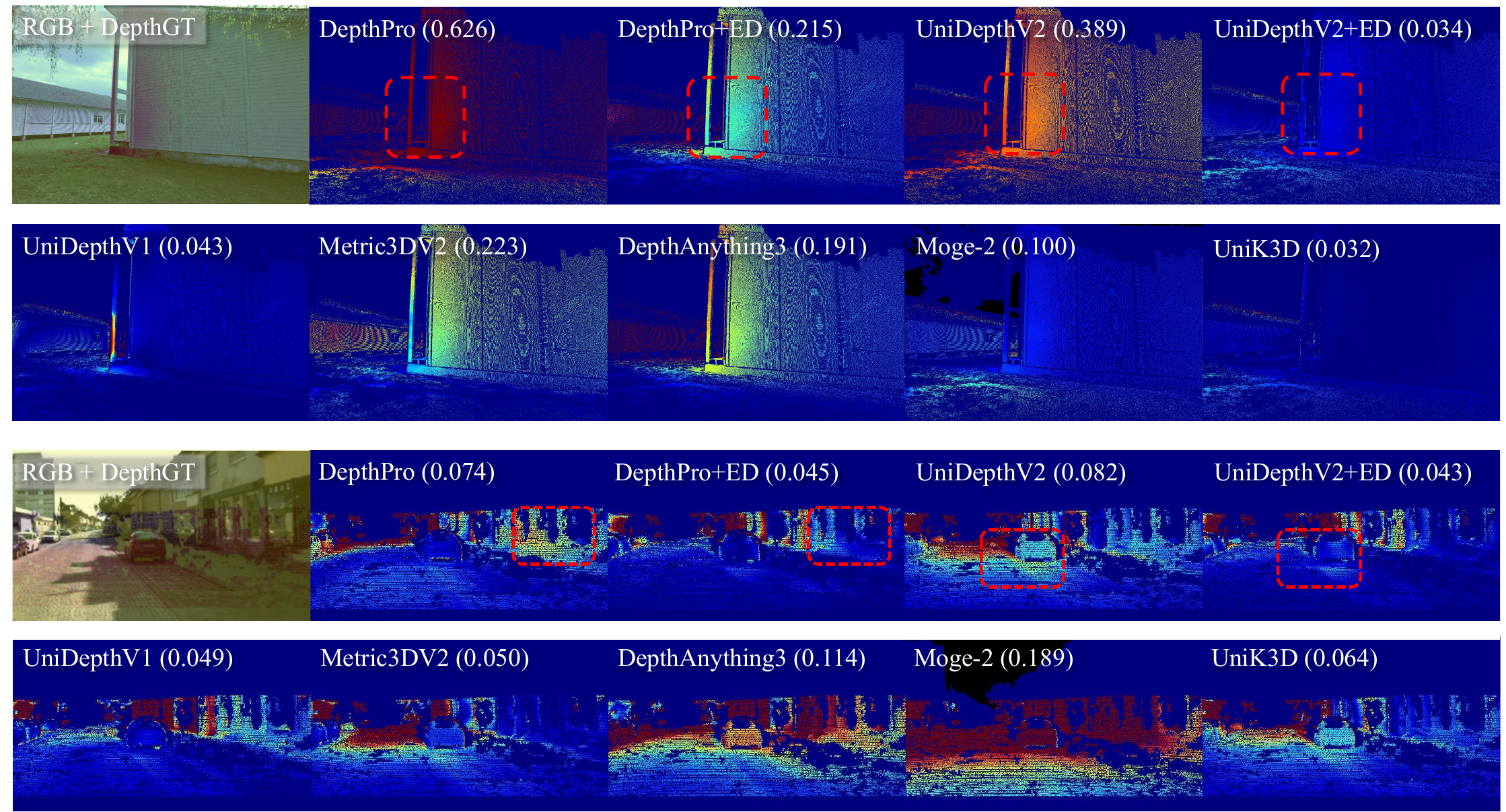} 
    \caption{\textbf{Qualitative comparison on the ETH3D \cite{eth3d:schops2017multi} (Top) and KITTI \cite{kitti:geiger2013vision} (Bottom) datasets.} The heatmaps visualize the absolute relative error (A.Rel), with the corresponding A.Rel value for each prediction reported in parentheses. ED indicates the proposed \textit{EpiDistill} method. Note that the KITTI visualizations have been cropped by 30\% on the left to improve visibility.}   \label{fig:qualitative}
\end{figure}

\subsection{Experimental Setup}

\textbf{Datasets.} We train our model on a multi-domain mixture of seven datasets covering indoor, outdoor, and synthetic environments: Hypersim \cite{hypersim:roberts2021hypersim}, Cityscapes \cite{cityscape:cordts2016cityscapes}, Eden \cite{eden:le2021eden}, ScanNet \cite{scannet:dai2017scannet}, ScanNet++ \cite{scannet++:yeshwanth2023scannet++}, Waymo \cite{waymo:sun2020scalability}, and nuScenes \cite{nuscenes:caesar2020nuscenes}. All selected datasets provide ground-truth depth, camera intrinsics, and 6-DoF camera poses. Using these annotations, we compute the visual overlap between frame pairs to construct training tuples consisting of a reference view and up to two source views, with triplets sampled within a $\pm$20 frame window.

To ensure that the epipolar attention can reliably establish meaningful geometric correspondences, we strictly enforce a minimum visual overlap of 20\% between the paired views. For the Hypersim dataset, identical sampling is applied using the exact ground-truth poses despite its discrete renderings. During training, input images are resized up to 576K pixels while preserving their aspect ratios, and the total number of training samples is capped at 50K per dataset to balance the training distribution.

\noindent \textbf{Evaluation Settings.} We evaluate depth prediction accuracy using standard metrics for monocular depth estimation: absolute relative error (Abs Rel), root mean squared error (RMSE), and $\delta_1 < 1.25$. Further details on these metrics can be found in the supplementary material. Crucially, our evaluation settings (\textit{e.g.}, depth clipping caps and valid region masking) strictly follow the evaluation protocol established by UniDepthV2 \cite{unidepthv2:piccinelli2025unidepthv2}.

\noindent \textbf{Training Details.} Our framework is built upon two representative Vision Transformer (ViT)-based monocular depth models: UniDepthV2 (ViT-L) \cite{unidepthv2:piccinelli2025unidepthv2} and DepthPro \cite{depthpro:bochkovskii2024depth}, utilizing their pre-trained weights. We intentionally select these two models to demonstrate the broad generalizability of \textit{EpiDistill} across different generations and performance tiers. Specifically, UniDepthV2 represents a cutting-edge SoTA approach, while DepthPro serves as a highly influential, established baseline. By consistently improving the performance of both models, we validate that our distillation method is not overfitted to a single architecture and can be seamlessly adapted to various ViT-based frameworks. 

For the architecture, we freeze the backbone of the base models to extract features, and introduce a lightweight DPT head \cite{dpt:ranftl2021vision} to predict dense shift offsets, alongside MLP-based camera and scale heads to regress the intrinsic residual and metric scale factor, respectively. Our epipolar distillation module configures 4 frame/epipolar layers with a $37 \times 37$ rectified stereo token grid per layer. 

During training, we apply random resize and center crop augmentations to simulate a diverse range of camera intrinsics. All models are trained for 100K iterations with a total batch size of 4 triplets (12 images), distributed across 8 NVIDIA H100 GPUs. The network is optimized using the AdamW \cite{AdamW:loshchilov2017decoupled} optimizer with a learning rate of $1 \times 10^{-4}$, momentum parameters $\beta_1=0.9$ and $\beta_2=0.999$, and a weight decay of $0.01$.

\begin{table}[t]
\centering
\caption{\textbf{Zero-shot outdoor and mixed dataset evaluation.} $\dagger$ indicates models using ground truth intrinsics during inference. Best results among models with only images as input are highlighted in \textbf{bold}, and second best are \underline{underlined}. The \textbf{Rank} is calculated across all 9 models.}
% 글자 밀림 방지 및 여백 확보를 위한 설정
\setlength{\tabcolsep}{5pt}
\renewcommand{\arraystretch}{1.1}
\resizebox{\textwidth}{!}{%
\begin{tabular}{l|cc|cc|cc|cc|c}
\toprule
\multirow{2}{*}{\textbf{Method/Metrics}} & \multicolumn{2}{c|}{\textbf{KITTI} \cite{kitti:geiger2013vision}} & \multicolumn{2}{c|}{\textbf{DDAD} \cite{ddad:guizilini20203d}} & \multicolumn{2}{c|}{\textbf{DIODE} \cite{diode:vasiljevic2019diode}} & \multicolumn{2}{c|}{\textbf{ETH3D} \cite{eth3d:schops2017multi}} & \multirow{2}{*}{\textbf{Rank} $\downarrow$} \\
 & A.Rel\,$\downarrow$ & $\delta_1$\,$\uparrow$  
 & A.Rel\,$\downarrow$ & $\delta_1$\,$\uparrow$  
 & A.Rel\,$\downarrow$ & $\delta_1$\,$\uparrow$  
 & A.Rel\,$\downarrow$ & $\delta_1$\,$\uparrow$ & \\
\midrule
Metric3DV2\textsuperscript{$\dagger$}\cite{metric3dv2:hu2024metric3d} & 0.054 & 0.975 & 0.122 & 0.858 & 0.536 & 0.077 & 0.175 & 0.724 & 4.5 \\
DepthAnything3\textsuperscript{$\dagger$}\cite{da3:lin2025depth} & 0.097 & 0.910 & 0.137 & 0.828 & 0.489 & 0.110 & 0.122 & 0.861 & 4.8 \\
\hhline{==========}  
UniDepthV1\cite{unidepth:piccinelli2024unidepth} & \textbf{0.049} & \textbf{0.980} & 0.217 & 0.708 & \textbf{0.263} & 0.624 & 0.530 & 0.224 & 4.8 \\
UniK3D\cite{unik3d:piccinelli2025unik3d} & 0.123 & 0.940 & 0.156 & \underline{0.859} & 0.457 & \textbf{0.686} & 0.139 & \underline{0.852} & \underline{4.0} \\
MoGe-2\cite{moge2:wangmoge} & 0.132 & 0.867 & 0.168 & 0.776 & 0.407 & 0.368 & \textbf{0.109} & \textbf{0.911} & 5.3 \\
\midrule
DepthPro\cite{depthpro:bochkovskii2024depth} & 0.141 & 0.838 & 0.391 & 0.233 & \underline{0.386} & 0.419 & 0.355 & 0.433 & 8.8 \\
+\textit{EpiDistill} & 0.123 & 0.875 & 0.348 & 0.353 & 0.357 & 0.462 & 0.354 & 0.393 & 6.4 \\
UniDepthV2\cite{unidepthv2:piccinelli2025unidepthv2} & 0.080 & 0.945 & \underline{0.144} & \textbf{0.859} & 0.716 & 0.549 & 0.176 & 0.752 & 4.7 \\
+\textit{EpiDistill} & \underline{0.074} & \underline{0.951} & \textbf{0.132} & 0.854 & 0.396 & \underline{0.641} & \underline{0.137} & 0.779 & \textbf{3.1} \\
\bottomrule
\end{tabular}
}
\label{tab:zeroshot_mixed}
\end{table}

\begin{table}[t]
\centering
 \caption{\textbf{Zero-shot indoor dataset evaluation.} $\dagger$ indicates models using ground truth intrinsics during inference. Best results among models with only images as input are highlighted in \textbf{bold}, and second best are \underline{underlined}. The \textbf{Rank} is calculated across all 9 models.}
% 글자 밀림 방지 및 여백 확보를 위한 설정
\setlength{\tabcolsep}{5pt}
\renewcommand{\arraystretch}{1.1}
\resizebox{\textwidth}{!}{%
\begin{tabular}{l|cc|cc|cc|cc|c}
\toprule
\multirow{2}{*}{\textbf{Method/Metrics}} & \multicolumn{2}{c|}{\textbf{NYU} \cite{nyu:silberman2012indoor}} & \multicolumn{2}{c|}{\textbf{Bonn} \cite{bonn:palazzolo2019iros}} & \multicolumn{2}{c|}{\textbf{Booster} \cite{boosterramirez2022open}} & \multicolumn{2}{c|}{\textbf{IBims-1}\cite{ibims:koch2020comparison}} & \multirow{2}{*}{\textbf{Rank} $\downarrow$} \\
 & A.Rel\,$\downarrow$ & $\delta_1$\,$\uparrow$  
 & A.Rel\,$\downarrow$ & $\delta_1$\,$\uparrow$  
 & A.Rel\,$\downarrow$ & $\delta_1$\,$\uparrow$  
 & A.Rel\,$\downarrow$ & $\delta_1$\,$\uparrow$ & \\
\midrule
Metric3DV2\textsuperscript{$\dagger$}\cite{metric3dv2:hu2024metric3d}  & 0.071 & 0.963 & 0.055 & 0.991 & 0.444 & 0.373 & 0.131 & 0.848 & 4.6 \\
DepthAnything3\textsuperscript{$\dagger$} \cite{da3:lin2025depth} & 0.078 & 0.957 & 0.048 & 0.991 & 0.178 & 0.771 & 0.079 & 0.961 & \textbf{2.4} \\
\hhline{==========}  
UniDepthV1\cite{unidepth:piccinelli2024unidepth} & \textbf{0.058} & \textbf{0.985} & 0.062 & 0.986 & 0.468 & 0.324 & 0.446 & 0.133 & 6.0 \\
UniK3D\cite{unik3d:piccinelli2025unik3d}  & 0.090 & 0.948 & 0.088 & 0.984  & 0.190 & 0.712 & 0.097 & 0.900 & 5.3 \\
MoGe-2\cite{moge2:wangmoge}  & 0.077 & 0.945 & 0.211 & 0.540 & 0.262 & 0.605 & 0.114 & 0.862 & 6.1 \\
\midrule
DepthPro\cite{depthpro:bochkovskii2024depth} & 0.098 & 0.918 & 0.091 & 0.912 & 0.332 & 0.535 & 0.156 & 0.837 & 7.8 \\
+\textit{EpiDistill} & 0.110 & 0.884  & 0.076 & 0.971 & 0.314 & 0.544 & 0.151 & 0.841 & 7.1 \\
UniDepthV2\cite{unidepthv2:piccinelli2025unidepthv2} & 0.077 & \underline{0.958} & \underline{0.057} & \underline{0.991} & \underline{0.179} & \underline{0.720} & \underline{0.093} & \underline{0.930} & 3.3 \\
+\textit{EpiDistill}  & \underline{0.077} & 0.958 & \textbf{0.055} & \textbf{0.991} & \textbf{0.171} & \textbf{0.727} & \textbf{0.092} & \textbf{0.935} & \textbf{2.4} \\
\bottomrule
\end{tabular}
}
\label{tab:zeroshot_indoor}
\end{table}

\subsection{Zero-Shot Depth Estimation}
We evaluate the zero-shot generalization of our metric depth estimation approach across diverse datasets. We compare our model against recent SoTA baselines, including Metric3Dv2 \cite{metric3dv2:hu2024metric3d}, UniDepthV1 \cite{unidepth:piccinelli2024unidepth}, UniK3D \cite{unik3d:piccinelli2025unik3d}, MoGe-2 \cite{moge2:wangmoge}, and DepthAnything3 (ViT-L, Metric) \cite{da3:lin2025depth}. Notably, Metric3DV2 and DepthAnything3 inherently rely on ground-truth (GT) camera intrinsics during inference, whereas other methods operate solely on images.

\noindent \textbf{Outdoor \& Mixed-Set Evaluation.} \cref{tab:zeroshot_mixed} evaluates geometric robustness across highly variable environments encompassing both indoor and outdoor scenes \cite{kitti:geiger2013vision, ddad:guizilini20203d, diode:vasiljevic2019diode, eth3d:schops2017multi}. Integrating \textit{EpiDistill} consistently elevates baseline performance across all domains. Gains are particularly prominent on the DIODE dataset, resulting in a 44.7\% gain in A.Rel metrics compared to UniDepthV2. Remarkably, UniDepthV2 + \textit{EpiDistill} achieves the best overall rank (3.1), outperforming even SoTA models reliant on GT intrinsics. As visually corroborated in highlighted regions of \cref{fig:qualitative}, \textit{EpiDistill} uniquely empowers both baselines to comprehensively resolve severe scale ambiguities and recover sharp structural details, yielding robust and globally consistent metric depth.

\noindent \textbf{Indoor Evaluation.} \cref{tab:zeroshot_indoor} reports quantitative results across four constrained indoor domains \cite{nyu:silberman2012indoor, bonn:palazzolo2019iros, boosterramirez2022open, ibims:koch2020comparison}. \textit{EpiDistill} brings stable improvements to both DepthPro \cite{depthpro:bochkovskii2024depth} and UniDepthV2 \cite{unidepthv2:piccinelli2025unidepthv2}. On the Booster dataset, our framework reduces the A.Rel error by 5.4\% and 4.5\% for DepthPro and UniDepthV2, respectively. Ultimately, UniDepthV2 + \textit{EpiDistill} achieves an average rank (2.4) matching the GT intrinsic-dependent DepthAnything3. While \textit{EpiDistill} primarily targets the severe scale ambiguities of unconstrained scenes, these results confirm that it simultaneously preserves and enhances accuracy in indoor environments where scale variations are less dominant.

\subsection{Generalization Across Diverse Scales}
To investigate our framework's robustness against extreme scale ambiguity, we evaluate scale generalization on the DepthPerturb dataset \cite{dollyzoom:nugent2025evaluating}, utilizing Dolly Zoom sequences. A Dolly Zoom effect occurs when the camera translates along the optical axis while simultaneously adjusting the focal length to keep the main subject's 2D size constant on the image plane. This configuration uniquely challenges a model's ability to disentangle metric depth from camera intrinsics, revealing whether it genuinely understands 3D geometry.

Since the dataset lacks ground truth camera information, we assess this disentanglement by reporting standard depth metrics alongside the Pearson Correlation (PCorr) for intrinsic evaluation. In a Dolly Zoom setup, the focal length is deliberately adjusted linearly to counteract the camera translation. Therefore, PCorr serves as a reliable indicator of the model's ability to capture this linear geometric trend. Formally, for a video of $T$ frames, let $t$ denote the frame index and $\hat{f}_t$ the predicted focal length. To measure temporal linearity, PCorr is computed as $|\text{PCorr}| = |\sum (t - \bar{t})(\hat{f}_t - \bar{\hat{f}})| / \sqrt{\sum (t - \bar{t})^2 \sum (\hat{f}_t - \bar{\hat{f}})^2}$, where $\bar{t}$ and $\bar{\hat{f}}$ are their respective means over the sequence.

\begin{wraptable}{r}{0.5\textwidth}
\centering
\vspace{-5mm}
\caption{Scale generalization evaluation on the DepthPerturb dataset \cite{dollyzoom:nugent2025evaluating}. \textit{EpiDistill} demonstrates superior disentanglement of metric depth and focal length.}
% 표가 지정한 폭(0.55\textwidth)을 넘어갈 경우를 대비해 resizebox 적용
\resizebox{\linewidth}{!}{
\begin{tabular}{lcccc}
\toprule
\multirow{2}{*}{\textbf{Method}} & \multicolumn{3}{c}{\textbf{Depth}} & \textbf{Intrinsic} \\
\cmidrule(lr){2-4} \cmidrule(l){5-5}
& A.Rel $\downarrow$ & RMS $\downarrow$ & $\delta_1 \uparrow$ & |PCorr| $\uparrow$ \\
\midrule
UniDepthV1 \cite{unidepth:piccinelli2024unidepth} & 0.432 & 0.957 & 0.350 & 0.704 \\
\midrule
DepthPro \cite{depthpro:bochkovskii2024depth} & 0.248 & 0.596 & 0.598 & 0.906 \\
\quad + \textit{EpiDistill} & 0.178 & \textbf{0.498} & 0.720 & 0.911 \\
\midrule
UniDepthV2 \cite{unidepthv2:piccinelli2025unidepthv2} & 0.241 & 0.853 & 0.689 & 0.955 \\
\quad + \textit{EpiDistill} & \textbf{0.167} &0.694 & \textbf{0.843} & \textbf{0.965} \\
\bottomrule
\end{tabular}
}
\label{tab:dollyzoom}
% \vspace{-3mm}
\end{wraptable}

As reported in \cref{tab:dollyzoom}, we evaluate three intrinsic-predicting baselines \cite{unidepth:piccinelli2024unidepth, depthpro:bochkovskii2024depth, unidepthv2:piccinelli2025unidepthv2}. Baseline models fail to perceive dynamic depth changes and struggle to predict the linearly varying intrinsics. In contrast, our \textit{EpiDistill} framework demonstrates exceptional generalization, significantly reducing depth errors (e.g., A.Rel drops from $0.241$ to $0.167$ for UniDepthV2, and $0.248$ to $0.178$ for DepthPro). Furthermore, it yields an improved focal length prediction trajectory, characterized by maximized PCorr scores. This confirms that transferring multi-view epipolar geometry equips the single-view model with the robust, intrinsic-aware scale priors necessary to resolve depth-intrinsic ambiguity.

\subsection{Ablation Study}
We conduct an ablation study to validate the individual and synergistic contributions of our proposed modules, namely Depth-Guided Epipolar Attention (DGEA) and Rectified Stereo Tokens (RST), using the UniDepthV2 model. To ensure a comprehensive evaluation, we report the averaged performance metrics across the KITTI \cite{kitti:geiger2013vision} and ETH3D \cite{eth3d:schops2017multi} datasets. The fine-tuned baseline refers to directly fine-tuning the decoder with a learning rate of $1\times10^{-6}$.

We first report the performance of the fine-tuned UniDepthV2 trained on our dataset. As an additional baseline, we employ the standard global cross-attention mechanism used in recent multi-view foundation models \cite{vggt:wang2025vggt, da3:lin2025depth}, which degenerates into intra-frame self-attention during single-view inference. Note that we omit an RST-only ablation variant, as the regularized sparse tokens inherently rely on the epipolar attention layer to establish and process geometric correspondences. Computational costs (FLOPs and latency) are measured on a single NVIDIA RTX A6000 GPU.

\noindent \textbf{Depth-Guided Epipolar Attention (DGEA).} The model is trained with the DGEA layer but inferred without RST. Consequently, during single-view inference, the reference image tokens serve simultaneously as query, key, and value, functioning similarly to standard frame attention. \cref{tab:ablation_components} shows consistent improvements across all metrics with negligible computational overhead. This indicates that DGEA effectively enforces explicit spatial constraints during multi-view training, embedding a geometric prior that indirectly benefits single-view inference compared to global attention.

\begin{table}[t]
    \centering
    \caption{Ablation study on the effectiveness of our proposed modules. Depth-Guided Epipolar Attention, Rectified Stereo Tokens. Best results are highlighted in \textbf{bold}.}
    \renewcommand{\arraystretch}{1.15} 
    \setlength{\tabcolsep}{10pt} 
    \resizebox{\textwidth}{!}{% 표 크기를 페이지 너비에 맞춤
    \begin{tabular}{cc cc ccc}
    \toprule
    \multicolumn{2}{c}{\textbf{Components}} &\multicolumn{2}{c}{\textbf{Computation Cost}} & \multicolumn{3}{c}{\textbf{Metrics}} \\
    \cmidrule(lr){1-2} \cmidrule(lr){3-4}  \cmidrule(lr){5-7}
    DGEA & RST & Memory & Inference time &A.Rel $\downarrow$ & RMS $\downarrow$ & $\delta_1 \uparrow$ \\
    \midrule
    \multicolumn{2}{l}{\textit{UniDepthV2 (Fine-tuned)}} &57.87M &131ms &0.124 &2.721  & 0.840\\    
    \multicolumn{2}{l}{\textit{UniDepthV2 (Global Attn.)}} &154.65M &180ms &0.119 &2.649  & 0.858\\
    \midrule 
    \checkmark &          &155.65M &190ms  &0.113  &2.419  &0.866  \\
               % & \checkmark & 0.256 & 2.777 & 0.683 \\
    \checkmark & \checkmark &180.98M &218ms &\textbf{0.101} &\textbf{ 2.206} & \textbf{0.870} \\
    \hhline{=======} 
    \multicolumn{2}{l}{\textit{Multi-view Model}} &180.98M &396ms &0.100  &2.102  &0.880 \\
        
    \bottomrule
    \end{tabular}
    }
    \vspace{-3mm}
    \label{tab:ablation_components}
\end{table}

\noindent \textbf{EpiDistill (Full Framework).} Integrating both DGEA and RST yields the most significant performance boost. RST preserves the cross-view attention pathways optimized by DGEA during multi-view training. By maintaining this continuous flow of geometry-aware information, our complete \textit{EpiDistill} framework explicitly mitigates scale ambiguity, achieving the highest $\delta_1$ accuracy and the lowest errors among single-view configurations, including a $16.7\%$ reduction in RMSE compared to the baseline. 

\noindent \textbf{Multi-view Upper Bound.} Finally, we evaluate the multi-view model to establish a performance upper bound. As shown in \cref{tab:ablation_components}, while the multi-view model requires identical memory, processing three input views nearly doubles the inference time to 396 ms. Notably, our single-view \textit{EpiDistill} closely approaches this multi-view upper bound with only marginal differences in overall metrics. This firmly demonstrates that \textit{EpiDistill} successfully distills multi-view geometric knowledge into a single-view predictor, effectively achieving multi-view-level accuracy without the computational overhead.

\section{Conclusion}
\label{sec:conclusion}

In this paper, we presented \textit{EpiDistill}, a novel geometric distillation framework designed to tackle the persistent challenge of scale ambiguity in monocular metric depth estimation. While recent multi-view foundation models exhibit robust scale perception, they inherently suffer from structural attention ambiguity when restricted to single-image inference. To bridge this gap, our approach transfers the geometry-aware scale priors from a multi-view to a single-view model. 

At the core of our methodology are two key innovations: the \textit{Depth-Guided Epipolar Attention}, which enforces strict spatial constraints during multi-view training, and \textit{Rectified Stereo Tokens}, which implicitly construct a geometric anchor to preserve cross-attention pathways during single-view inference. Extensive zero-shot evaluations demonstrate that our model-agnostic approach significantly enhances the performance of SoTA ViT-based baselines. By effectively decoupling and retaining multi-view scale knowledge, \textit{EpiDistill} exhibits remarkable robustness across diverse outdoor and indoor environments, and proves particularly resilient against extreme depth-intrinsic entanglements, as evidenced by our scale generalization analysis.

% \newpage
% ---- Bibliography ----
%
% BibTeX users should specify bibliography style 'splncs04'.
% References will then be sorted and formatted in the correct style.
%
% \bibliographystyle{splncs04}
% \bibliography{main}

\bibliographystyle{splncs04}
% \bibliography{main} % <--- (기존 코드) 지우거나 주석 처리
\bibliography{main} % <--- (수정) 쉼표(,)로 두 bib 파일을 연결

\end{document}

% ---------------------------------------------------------------
% TODO REVIEW: Replace with your title
% 수정된 코드:
\title{Geometric Distillation from Rectified Stereo: Leveraging Epipolar Cues for Monocular Depth}

% TODO REVIEW: If the paper title is too long for the running head, you can set
% an abbreviated paper title here. If not, comment out.
\titlerunning{EpiDistill}

% TODO FINAL: Replace with your author list. 
% Include the authors' OCRID for the camera-ready version, if at all possible.
% \author{Jung-Hee Kim\inst{1}\orcidlink{0009-0007-3198-3029} \and
% Xiaoming Liu\inst{1,2}\orcidlink{0000-0003-3215-8753}}

% % TODO FINAL: Replace with an abbreviated list of authors.
% \authorrunning{J.~Kim and X.~Liu}
% % First names are abbreviated in the running head.
% % If there are more than two authors, 'et al.' is used.

% % TODO FINAL: Replace with your institution list.
% \institute{Michigan State University, East Lansing, MI 48824, USA \\ \and 
% University of North Carolina at Chapel Hill, Chapel Hill, NC 27514, USA \\
% \email{kimjun84@msu.edu, liuxm@cs.unc.edu}}
  
\author{\quad} % 시각적으로 보이지 않는 띄어쓰기 기호 입력
\institute{\vspace{-4em}}
  
\makesupptitle

\section{Additional Analysis}
\begin{center}
    \includegraphics[width=1.0\linewidth]{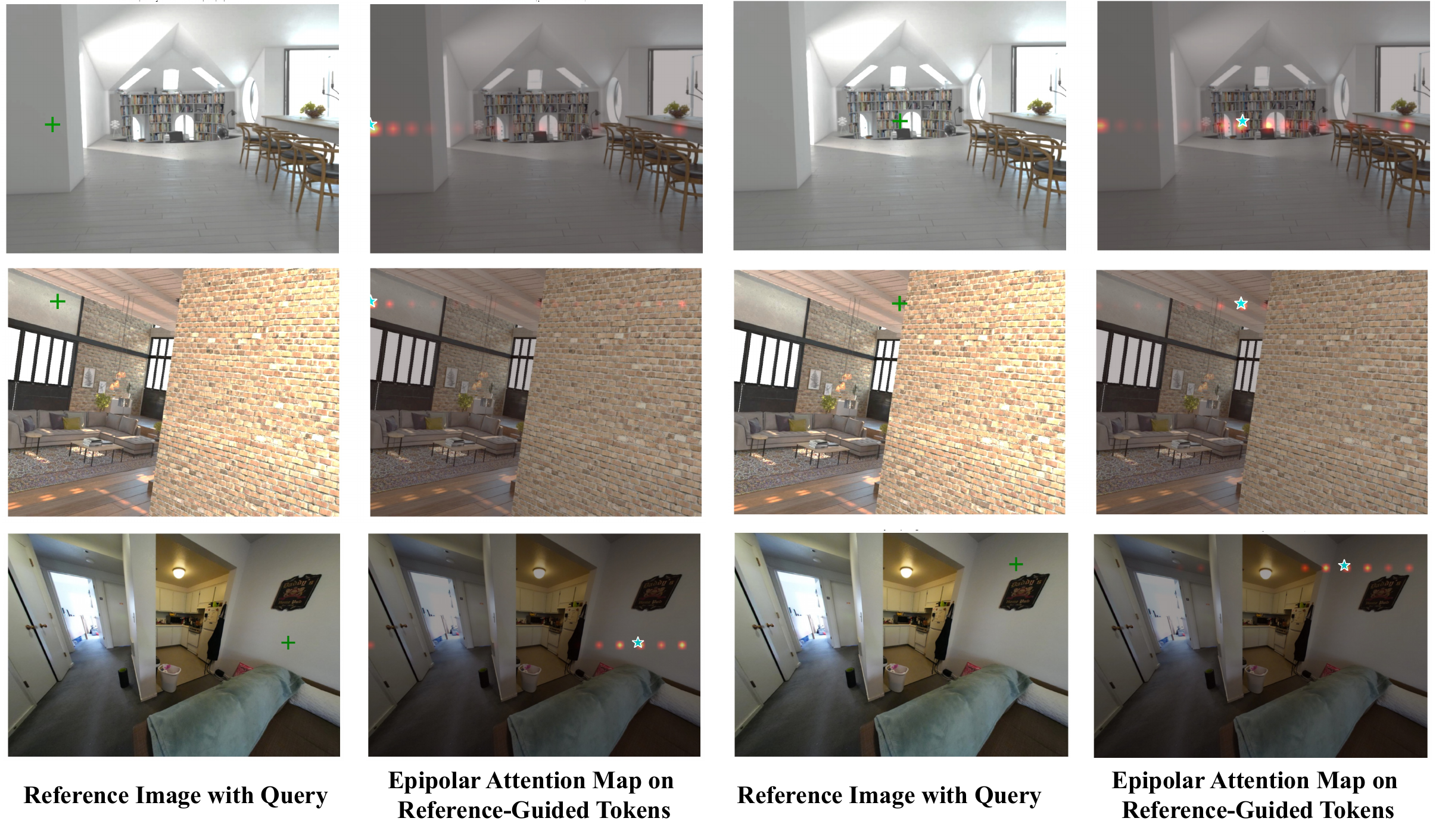} 
    \captionof{figure}{\textbf{Visualization of Epipolar Attention Maps.} For a given query point (green cross) in the reference image, we visualize the corresponding cross-attention weights (red heatmap) from the reference-guided tokens ($\mathbf{T}_{\text{rst-ref}}$). The cyan star indicates the location of peak attention. Note that the attention distributions are strictly constrained along the horizontal epipolar lines under the rectified stereo condition.}   
    \label{fig:supp_analysis1}
\end{center}

\subsection{Epipolar Attention Visualization}
To provide deeper insights into our geometric guidance mechanism, we visualize the attention map between the reference image features ($\mathbf{F}_{\text{ref}}$) and the reference-guided tokens ($\mathbf{T}_{\text{rst-ref}}$). Specifically, \cref{fig:supp_analysis1} displays the attention weights for selected query pixels to reference-guided tokens overlaid on the reference image, to illustrate how the network searches along the rectified epipolar line. This demonstrates that the reference-guided tokens successfully maintain valid epipolar geometry under a rectified stereo setup.

The visualizations demonstrate that the attention mechanism successfully forms highly localized peaks (cyan stars) along the horizontal epipolar lines. This behavior confirms that our rectified stereo tokens ($\mathbf{T}_{\text{rst}}$), combined with cross-attention, effectively serve as geometric anchors, establishing a distinct, focused attention peak along the epipolar line. Importantly, the spatial offset (\textit{i.e.}, disparity) between the query point and the attention peak strongly correlates with the underlying 3D scene geometry, indicating that the network has learned an implicit virtual baseline. As clearly seen in the figure, query points located on distant objects exhibit minimal disparity. Conversely, queries placed on nearby foreground objects produce larger disparities. This indicates that \textit{EpiDistill} successfully learns to encode and utilize multi-view geometric principles and transfers them to a rectified stereo setup.

\subsection{Dataset-Specific Adaptation of Rectified Stereo Tokens}

\begin{table}[t]
\centering
\caption{\textbf{Effect of Rectified Stereo Token Adaptation.} Fine-tuning the rectified stereo tokens and the cross-attention layer on specific datasets (DIODE\cite{diode:vasiljevic2019diode}, KITTI\cite{kitti:geiger2013vision}) yields consistent performance improvements. This proves that our tokens not only serve as geometric anchors but also act as highly efficient parameters for adapting to dataset-specific depth and camera priors.}
\label{tab:token_adaptation}
\begin{tabular}{@{\hspace{2mm}} l@{\hspace{3mm}}l @{\hspace{2mm}} c @{\hspace{2mm}}c @{\hspace{2mm}} c @{\hspace{2mm}}c @{\hspace{2mm}}}
\toprule
\multirow{2}{*}{\textbf{Model}} & \multirow{2}{*}{\textbf{Setting}} & \multicolumn{2}{c}{\textbf{DIODE} \cite{diode:vasiljevic2019diode}} & \multicolumn{2}{c}{\textbf{KITTI \cite{kitti:geiger2013vision}}} \\
% 밑줄도 중앙 정렬에 맞게 좌우(lr)를 살짝 깎아줍니다.
\cmidrule(lr){3-4} \cmidrule(lr){5-6}
& & A.Rel\,$\downarrow$ & $\delta_1$\,$\uparrow$ & A.Rel\,$\downarrow$ & $\delta_1$\,$\uparrow$ \\
\midrule
\multirow{2}{*}{DepthPro\cite{depthpro:bochkovskii2024depth}+ED} & Zero-shot & 0.357 & 0.462 & 0.123 & 0.875 \\
& Fine-tuned & \textbf{0.348}& \textbf{0.482} & \textbf{0.115} & \textbf{0.898} \\
\midrule
\multirow{2}{*}{UniDepthV2\cite{unidepthv2:piccinelli2025unidepthv2}+ED} & Zero-shot & 0.396  & 0.641 & 0.074  & 0.951 \\
& Fine-tuned & \textbf{0.380} & \textbf{0.663}  & \textbf{0.068} & \textbf{0.962} \\
\bottomrule
\end{tabular}%
% }
\end{table}

To investigate the adaptability of our geometric formulation, we conduct an ablation experiment where we exclusively fine-tune the rectified stereo tokens ($\mathbf{T}_{\text{rst}}$) and the cross-attention layer responsible for forming the reference-guided tokens ($\mathbf{T}_{\text{rst-ref}}$). Direct depth supervision is applied while the rest of the network remains strictly frozen. This lightweight adaptation is performed with a learning rate of $1 \times 10^{-6}$ and a batch size of 8 for only 3 epochs. We fine-tune the tokens on the training splits of the DIODE~\cite{diode:vasiljevic2019diode} and KITTI~\cite{kitti:geiger2013vision} datasets and evaluate them on their respective test sets.

As reported in \cref{tab:token_adaptation}, this dataset-specific fine-tuning yields consistent performance improvements over the zero-shot baselines. For instance, fine-tuning reduces the absolute relative error (A.Rel) by up to 8.1\% on KITTI and 4.0\% on DIODE datasets when applying \textit{EpiDistill} upon the UniDepthV2 model. This observation demonstrates a crucial property of our method: while the rectified stereo tokens act as robust domain-agnostic geometric anchors in zero-shot settings, they also serve as highly efficient parameters for dataset calibration. By updating these tokens, the network can seamlessly adapt to the specific camera and depth distributions of a target domain, circumventing the need for computationally expensive full-model fine-tuning.

\section{Additional Experiments}
\subsection{Additional Qualitative Results}
\begin{figure}[htbp]
    \centering
    \includegraphics[width=1.0\linewidth]{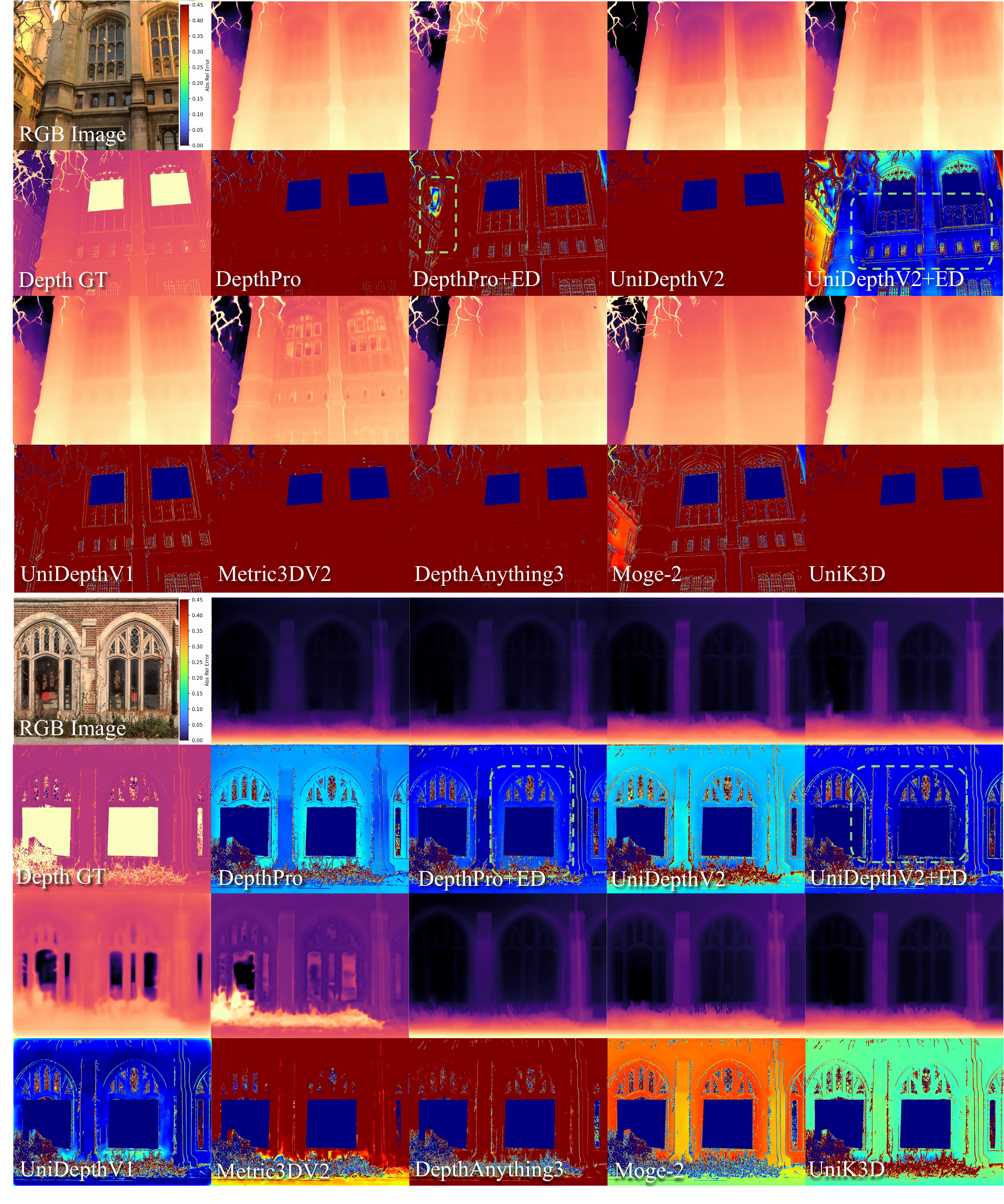} 
    \caption{\textbf{Qualitative comparison on the DIODE dataset.} Visualizations of predicted depth maps and their corresponding absolute relative error (A.Rel) heatmaps. `ED' denotes the proposed \textit{EpiDistill} method. The green dashed boxes highlight regions where the metric scale is significantly corrected.}   
    \label{fig:supp_qual1}
\end{figure}

\begin{figure}[htbp]
    \centering
    \includegraphics[width=1.0\linewidth]{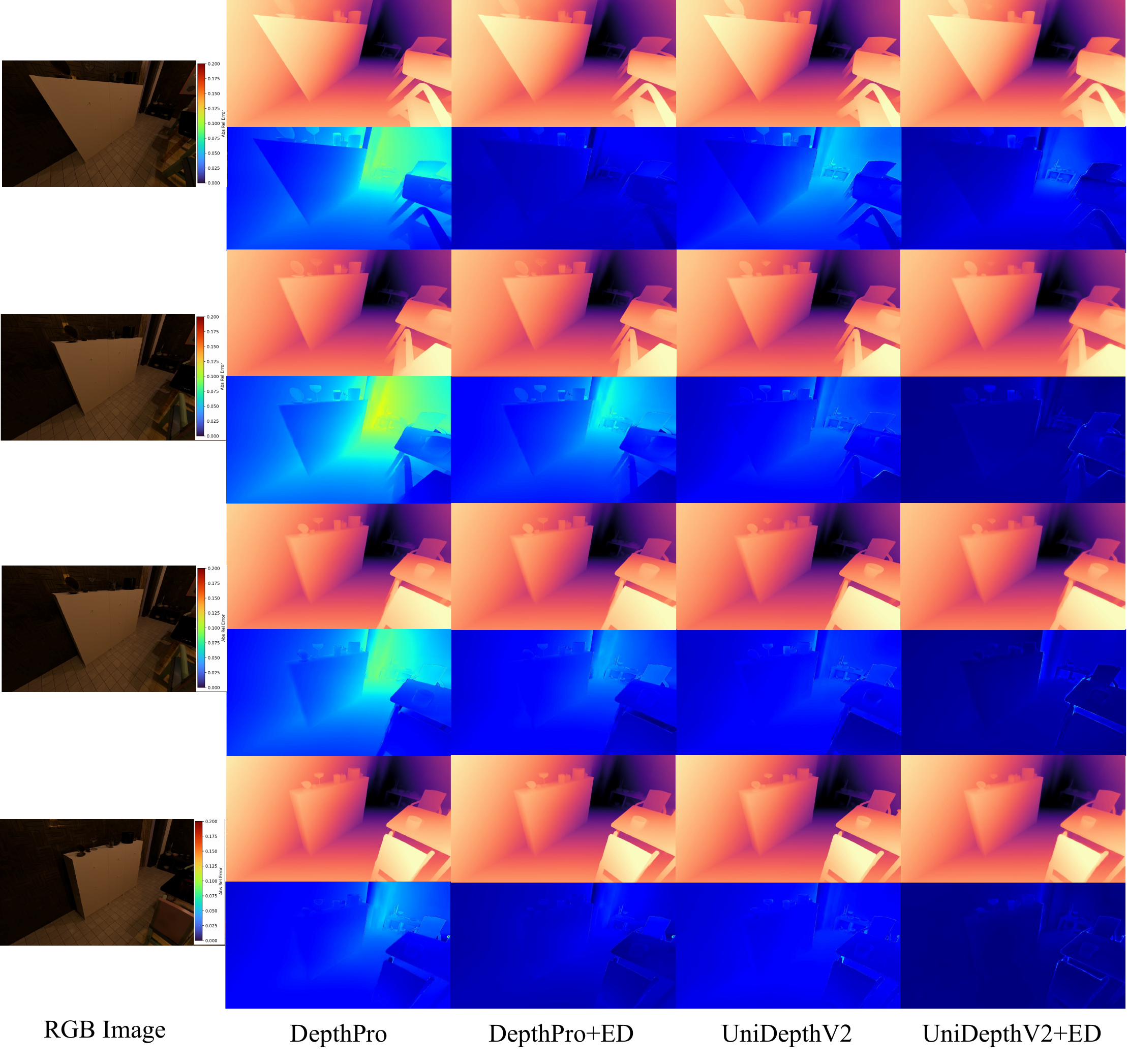} 
    \caption{\textbf{Qualitative comparison on a zoom sequence from the DepthPerturb dataset.} The rows represent a dynamic sequence with continuously varying focal lengths, where the column indicates the depth map and A.Rel error map of each model. While baselines struggle to maintain consistent scale, our \textit{EpiDistill} (+ED) yields consistently accurate metric predictions across all focal lengths.}   
    \label{fig:supp_qual2}
\end{figure}
We provide additional qualitative comparisons on the DIODE~\cite{diode:vasiljevic2019diode} and DepthPerturb~\cite{dollyzoom:nugent2025evaluating} datasets to highlight the significant scale improvements achieved by our method. As shown in \cref{fig:supp_qual1}, state-of-the-art (SoTA) baselines inherently suffer from scale ambiguity when predicting metric depth, resulting in large absolute relative errors, while maintaining structure information as shown in the depth map results. Integrating our proposed \textit{EpiDistill} (+ED) effectively resolves this limitation, correctly restoring the accurate scale. Furthermore, \cref{fig:supp_qual2} demonstrates our model's robustness to dynamic camera intrinsic changes using a dolly-zoom sequence. While the continuously varying focal length causes baseline models to produce fluctuating and inconsistent scale predictions, models augmented with \textit{EpiDistill} consistently deliver accurate and stable metric scale estimation. Together, these results prove that our geometric guidance robustly recovers scale even under challenging and dynamic camera configurations.

\subsection{Scale vs. Relative Depth Comparison}

\begin{table*}[t]
\centering
\caption{\textbf{Impact of EpiDistill on Metric vs. Relative Depth across Datasets.} We report the A.Rel ($\downarrow$) for zero-shot evaluation on KITTI, DIODE, and Booster. Integrating \textit{EpiDistill} yields large improvements in metric scale across all diverse datasets, while preserving or slightly improving the already strong relative structural predictions of the baselines.}
\label{tab:supp_scale_improvement}
% \resizebox{\textwidth}{!}{%
\begin{tabular}{l @{\hspace{3mm}} cc @{\hspace{3mm}} cc @{\hspace{3mm}} cc}
\toprule
\multirow{2}{*}{\textbf{Method}} & \multicolumn{2}{c}{\textbf{KITTI\cite{kitti:geiger2013vision}}} & \multicolumn{2}{c}{\textbf{DIODE\cite{diode:vasiljevic2019diode}}} & \multicolumn{2}{c}{\textbf{Booster\cite{boosterramirez2022open}}} \\
\cmidrule(r{1mm}){2-3} \cmidrule(r{1mm}){4-5} \cmidrule{6-7}
 & \textbf{Metric} & \textbf{Relative} & \textbf{Metric} & \textbf{Relative} & \textbf{Metric} & \textbf{Relative} \\
\midrule
DepthPro\cite{depthpro:bochkovskii2024depth} & 0.141 & 0.085 & 0.386 & 0.438 & 0.332 & 0.021 \\
\textbf{+ \textit{EpiDistill}} & \textbf{0.123} & \textbf{0.076} & \textbf{0.357} & \textbf{0.430} & \textbf{0.314} & \textbf{0.021} \\
\rowcolor{gray!15} \textit{Improvement} & \textit{\textbf{+12.8\%}} & \textit{+10.6\%} & \textit{\textbf{+7.5\%}} & \textit{+1.8\%} & \textit{\textbf{+5.4\%}} & \textit{0.0\%} \\
\midrule
UniDepthV2\cite{unidepthv2:piccinelli2025unidepthv2} & 0.080 & 0.056 & 0.716 & 0.427 & 0.179 & 0.029 \\
\textbf{+ \textit{EpiDistill}} & \textbf{0.074} & \textbf{0.056} & \textbf{0.396} & \textbf{0.368} & \textbf{0.171} & 0.033 \\
\rowcolor{gray!15} \textit{Improvement} & \textit{\textbf{+7.5\%}} & \textit{+0.1\%} & \textit{\textbf{+44.7\%}} & \textit{+13.8\%} & \textit{\textbf{+4.5\%}} & \textit{-12.1\%} \\
\bottomrule
\end{tabular}%
% }
\end{table*}

To explicitly demonstrate that the performance gains achieved by \textit{EpiDistill} stem primarily from resolving scale ambiguity, we evaluate the depth estimation accuracy under both metric and relative settings. 

As reported in \cref{tab:supp_scale_improvement}, baselines demonstrate strong performance when predictions are scale-aligned to the ground truth (Relative Depth) via median scaling. By integrating \textit{EpiDistill}, we observe large improvements on metric depth evaluation criteria (A.Rel). Conversely, the performance margins in the relative depth setting remain marginal or comparable to the baselines. These results confirm our hypothesis: the structural integrity of the baseline models is already robust, and the primary contribution of \textit{EpiDistill} lies in its ability to effectively inject accurate geometric scale cues, bridging the gap between relative and metric depth estimation.

\begin{table}[t]
    \centering
    \caption{Additional ablation studies on loss function variants, Regularized Sparse Token (RST) resolutions, and hyperparameters of Depth-Guided Epipolar Attention (DGEA) Gaussians.}
    \label{tab:add_ablation_study}
    \small 
    \setlength{\tabcolsep}{6pt} 
    \begin{tabularx}{\columnwidth}{l|X|cc} % 우측 메트릭 사이의 세로선(|)을 빼는 것이 훨씬 깔끔합니다.
        \toprule
        \textbf{Module} & \textbf{Experiments} & \textbf{Abs Rel} ($\downarrow$) & $\boldsymbol{\delta_1}$ ($\uparrow$) \\  
        \midrule
        \multirow{3}{*}{Loss components}  
        & $\mathcal{L}_{\text{rel}} + \mathcal{L}_{\text{scale}} + \mathcal{L}_{\text{ray}}$ & 0.115 & 0.857 \\
        & $+ \mathcal{L}_{\text{attn}}$ only & 0.111 & 0.862 \\
        & $+ \mathcal{L}_{\text{distill}}$ only & 0.107 & 0.863 \\
        \midrule 
        \multirow{2}{*}{RST resolution} 
        & $33 \times 33$ Tokens & 0.103 & 0.867 \\ 
        & $37 \times 41$ Tokens & 0.101 & 0.869 \\
        \midrule 
        \multirow{3}{*}{DGEA Gaussians}
        & Width=8, Bias=50 & 0.105 & 0.864 \\
        & Width=16, Bias=25 & 0.102 & 0.867 \\       
        & Width=16, Bias=100 & 0.114 & 0.852 \\
        \midrule 
        \midrule
        \rowcolor{gray!10} 
        Ours (Full) & $37 \times 37$ Tokens, Width=16, Bias=50 & \textbf{0.101} & \textbf{0.870} \\
        \bottomrule
    \end{tabularx}
\end{table}

\subsection{Additional Ablation Studies}
\cref{tab:add_ablation_study} reports performance metrics following the identical evaluation protocol used in Table 4 of the main paper. We first analyze the individual impact of each loss component. Starting from a baseline objective consisting of relative, scale, and ray losses for intrinsic and depth estimation following recent literature \cite{unidepthv2:piccinelli2025unidepthv2, moge2:wangmoge}, we incrementally introduce our attention guidance ($\mathcal{L}_{\text{attn}}$) and token distillation ($\mathcal{L}_{\text{distill}}$) losses. The results demonstrate that both components consistently and incrementally improve estimation accuracy.

Furthermore, we ablate the configurations of the RST resolution and DGEA Gaussian parameters (Width and Bias). We observe that variations in token resolution yield marginal differences in depth accuracy; thus, we maintain our default resolution of $37 \times 37$ to balance performance and efficiency. Similarly, adjusting the Gaussian width does not significantly alter the results. However, applying an excessively high bias parameter (\textit{i.e.}, Bias=100) severely sharpens the attention map, leading to performance degradation. This is primarily because an overly peaked attention distribution forces the model to overfit to localized regions. In the presence of occlusions or dynamic objects across different views, such rigid supervision inherently introduces noisy attention targets, causing the distillation process to misalign. Ultimately, our proposed configuration ($\text{Bias}=50$) achieves the optimal trade-off, yielding robust and substantial accuracy improvements in A.Rel over all alternative configurations.

\section{Implementation Details}
\subsection{Model Pipeline}
The overall pipeline consists of a frozen transformer encoder for spatial feature extraction, frame and epipolar attention layers, a lightweight DPT decoder, a camera head, and a scale prediction head. 

We initialize a set of learnable tokens, including scale tokens and rectified stereo tokens. The spatial tokens obtained from the pretrained foundation encoders are concatenated with their respective class tokens. We then append the scale tokens to each viewpoint's token sequence. These sequences are processed through frame attention and depth-guided epipolar attention. During the epipolar attention phase, tokens cross-attend to source view features to align cross-view geometry according to relative camera poses. To sample along the continuous epipolar line, we uniformly extract 16 points.

Given the input reference and source images, the frozen backbone extracts multiscale spatial features with an output dimension of 1024. Note that to align the multi-level feature dimensions of DepthPro~\cite{depthpro:bochkovskii2024depth} (256, 512, 1024), we employ a simple linear projection using a single MLP layer. After the tokens are refined through the epipolar attention module, the attended spatial tokens are passed into the lightweight Dense Prediction Transformer (DPT) \cite{dpt:ranftl2021vision} decoder. This decoder produces a structural offset $\Delta d$. Simultaneously, the focal tokens are processed through 3-layer MLPs within the scale and camera heads to predict the global metric scale factor ($\hat{s}$) and camera intrinsic residuals ($\delta_x, \delta_y$), respectively.

The structural offset $\Delta d$ and the metric scale factor $\hat{s}$ are explicitly decoupled during decoding. The final absolute metric depth $\mathbf{D}_{\text{adj}}$ is assembled via pixel-wise multiplication:
\begin{equation}
    \mathbf{D}_{\text{adj}} = \hat{s} \cdot (\mathbf{D}_{\text{base}} + \Delta d).
\end{equation}
Additionally, the adjusted focal length is computed as $f_{\text{adj}} = f_{\text{base}} + (\delta_x W, \delta_y H)$, where $W$ and $H$ indicate the width and height of the image, as described in the main paper.

\subsection{Depth Evaluation Metrics}
Let $\mathbf{D}_{\text{GT}}$ and $\mathbf{\hat{D}}$ denote the ground truth and predicted depth maps, respectively, and $\Omega$ be the set of valid pixels. We evaluate our method using the following standard metrics:
\begin{align*}
    \text{A.Rel} &= \frac{1}{|\Omega|} \sum_{p \in \Omega} \frac{\bigl|\mathbf{\hat{D}}_p - \mathbf{D}_{\text{GT}, p}\bigr|}{\mathbf{D}_{\text{GT}, p}}, \quad
    \text{RMS} = \sqrt{\frac{1}{|\Omega|} \sum_{p \in \Omega} \bigl(\mathbf{\hat{D}}_p - \mathbf{D}_{\text{GT}, p}\bigr)^2}, \\
    \delta_1 &= \frac{1}{|\Omega|} \sum_{p \in \Omega} \mathbf{1}\!\Bigl( \max\!\Bigl(\frac{\mathbf{\hat{D}}_p}{\mathbf{D}_{\text{GT}, p}}, \frac{\mathbf{D}_{\text{GT}, p}}{\mathbf{\hat{D}}_p}\Bigr) < 1.25 \Bigr).
\end{align*}

% ---- Bibliography ----
%
% BibTeX users should specify bibliography style 'splncs04'.
% References will then be sorted and formatted in the correct style.
%
\bibliographystyle{splncs04}
\bibliography{supp}